\documentclass[10pt,twocolumn,letterpaper]{article}

\usepackage{iccv}              

\definecolor{iccvblue}{rgb}{0.21,0.49,0.74}
\usepackage[pagebackref,breaklinks,colorlinks,allcolors=iccvblue]{hyperref}

\usepackage[utf8]{inputenc} 
\usepackage[T1]{fontenc}    
\usepackage{url}            
\usepackage{booktabs}       
\usepackage{amsfonts}       
\usepackage{nicefrac}       
\usepackage{microtype}      
\usepackage{xcolor}         
\usepackage{colortbl}
\usepackage{comment}
\usepackage{color}
\usepackage{graphicx}
\usepackage{algpseudocode}
\usepackage{threeparttable}
\usepackage{multirow}
\usepackage{pifont}
\usepackage{makecell}
\usepackage{caption,subcaption}
\usepackage{dblfloatfix}

\usepackage[ruled,vlined]{algorithm2e}

\def\paperID{21} 
\def\confName{ICCV}
\def\confYear{2025}

\title{Attn-Adapter: Attention Is All You Need for \\
Online Few-shot Learner of Vision-Language Model}

\author{
Phuoc-Nguyen Bui\thanks{Equal contribution}\\
Sungkyunkwan University, Suwon, Korea\\
\and
Khanh-Binh Nguyen\footnotemark[1]\\
Deakin University, Geelong, Australia\\
\and
Hyunseung Choo\thanks{Corresponding author}\\
Sungkyunkwan University, Suwon, Korea\\
}

\begin{document}
\maketitle
\begin{abstract}
Contrastive vision-language models excel in zero-shot image recognition but face challenges in few-shot scenarios due to computationally intensive offline fine-tuning using prompt learning, which risks overfitting. To overcome these limitations, we propose Attn-Adapter, a novel online few-shot learning framework that enhances CLIP’s adaptability via a dual attention mechanism. Our design incorporates dataset-specific information through two components: the Memory Attn-Adapter, which refines category embeddings using support examples, and the Local-Global Attn-Adapter, which enriches image embeddings by integrating local and global features. This architecture enables dynamic adaptation from a few labeled samples without retraining the base model. Attn-Adapter outperforms state-of-the-art methods in cross-category and cross-dataset generalization, maintaining efficient inference and scaling across CLIP backbones.
\end{abstract}    
\section{Introduction}
\label{sec:intro}

Vision-language models (VLMs) unify visual and textual understanding for multimodal tasks. CLIP~\cite{radford2021learning}, a prominent example, enables zero-shot image recognition via large-scale contrastive learning between images and text. This allows CLIP to generalize across diverse visual concepts without category-specific supervision. However, many real-world tasks in medical imaging or robotics, require domain adaptation with limited labeled data \cite{nguyen2023boosting,nguyen2024debiasing,nguyen2024sequencematch}. Few-shot learning addresses this by adapting models to novel classes with minimal supervision. Offline methods like CoOp~\cite{zhou2022learning}, CoCoOp~\cite{zhou2022conditional}, ProMIM~\cite{bui2025accelerating}, and CLIP-Adapter~\cite{gao2021clip} fine-tune prompts or models using support data, but they are compute intensive and prone to overfitting. Online methods such as Tip-Adapter~\cite{zhang2021tip}, Proto-CLIP~\cite{palanisamy2024proto}, and Meta-Adapter~\cite{song2023meta} avoid full fine-tuning by via support features. Meta-Adapter improves generalization using a lightweight residual adapter, but still struggles to capture dataset-specific nuances due to reliance on zero-shot CLIP features \cite{nguyen2025calibration}.

To address these challenges, we propose Attn-Adapter, a novel online few-shot learner for vision-language models that leverages attention mechanisms to dynamically refine both category and image embeddings. Our approach introduces two key components: (1) a Memory Attn-Adapter that applies cross-attention to refine category embeddings using support embeddings as keys and values, and (2) a Local-Global Attn-Adapter that enhances image embeddings by integrating local and global features through attention mechanisms. Unlike previous methods that only fine-tune the affinity matrix from few-shot support samples, Attn-Adapter imposes dataset-specific information during fine-tuning through its dual attention mechanism, enabling more effective generalization across diverse datasets and tasks. Our contributions can be summarized as follows:

\begin{itemize}
    \item We propose Attn-Adapter, a lightweight online few-shot learner that leverages attention mechanisms to dynamically refine CLIP features guided by few-shot samples.
    \item We introduce a novel dual attention architecture consisting of Memory Attn-Adapter and Local-Global Attn-Adapter, which effectively captures dataset-specific information and enhances both category and image embeddings.
    \item Extensive experiments show that Attn-Adapter outperforms SOTA online methods across different configurations while maintaining higher inference speed.
\end{itemize}

\section{Related Work}

\subsection{Few-Shot Learning}

\paragraph{Offline methods}
Prompt learning is a popular offline strategy. CoOp~\cite{zhou2022learning} optimizes learnable prompt vectors for CLIP, improving dataset-specific performance. CoCoOp~\cite{zhou2022conditional} enhances generalization by conditioning prompts on image features. CLIP-Adapter~\cite{gao2021clip} uses lightweight adapters for feature-level adaptation. However, these methods are computationally expensive, prone to overfitting, and require retraining for new tasks, limiting their practicality. 

\paragraph{Online methods}
Online methods ~\cite{zhang2021tip, song2023meta, palanisamy2024proto, nguyen2025adaptive} enable adaptation without backbone updates. Tip-Adapter~\cite{zhang2021tip} uses a training-free cache-based approach for efficient inference but requires dataset-specific hyperparameter tuning. Meta-Adapter~\cite{song2023meta} improves robustness with gated attention trained through meta-learning, though it relies on static CLIP features. Proto-CLIP~\cite{palanisamy2024proto} aligns image-text prototypes but lacks flexibility due to per-dataset optimization.

\subsection{Attention Mechanisms in VLMs}
Attention mechanisms are central to VLMs, enabling focus on relevant input features. They operate via self-attention within modalities and cross-attention across modalities. Transformer-based VLMs, like CLIP, use multi-head self-attention to model relationships within image patches or text tokens. Cross-attention, as in LXMERT~\cite{tan2019lxmert} and ViLBERT~\cite{lu2019vilbert}, aligns visual and textual features for multimodal reasoning. Recent advances, such as Gallop~\cite{lafon2025gallop}, enhance few-shot learning by using attention to integrate local and global visual features in prompt learning. Our proposed Attn-Adapter leverages attention to dynamically refine category and image embeddings with dataset-specific cues for online few-shot learning.
\section{Methodology}
\label{sec:formatting}
\begin{figure*}[h]
    \centering
    \includegraphics[width=\textwidth, trim=0mm 0mm 8mm 0mm, clip]{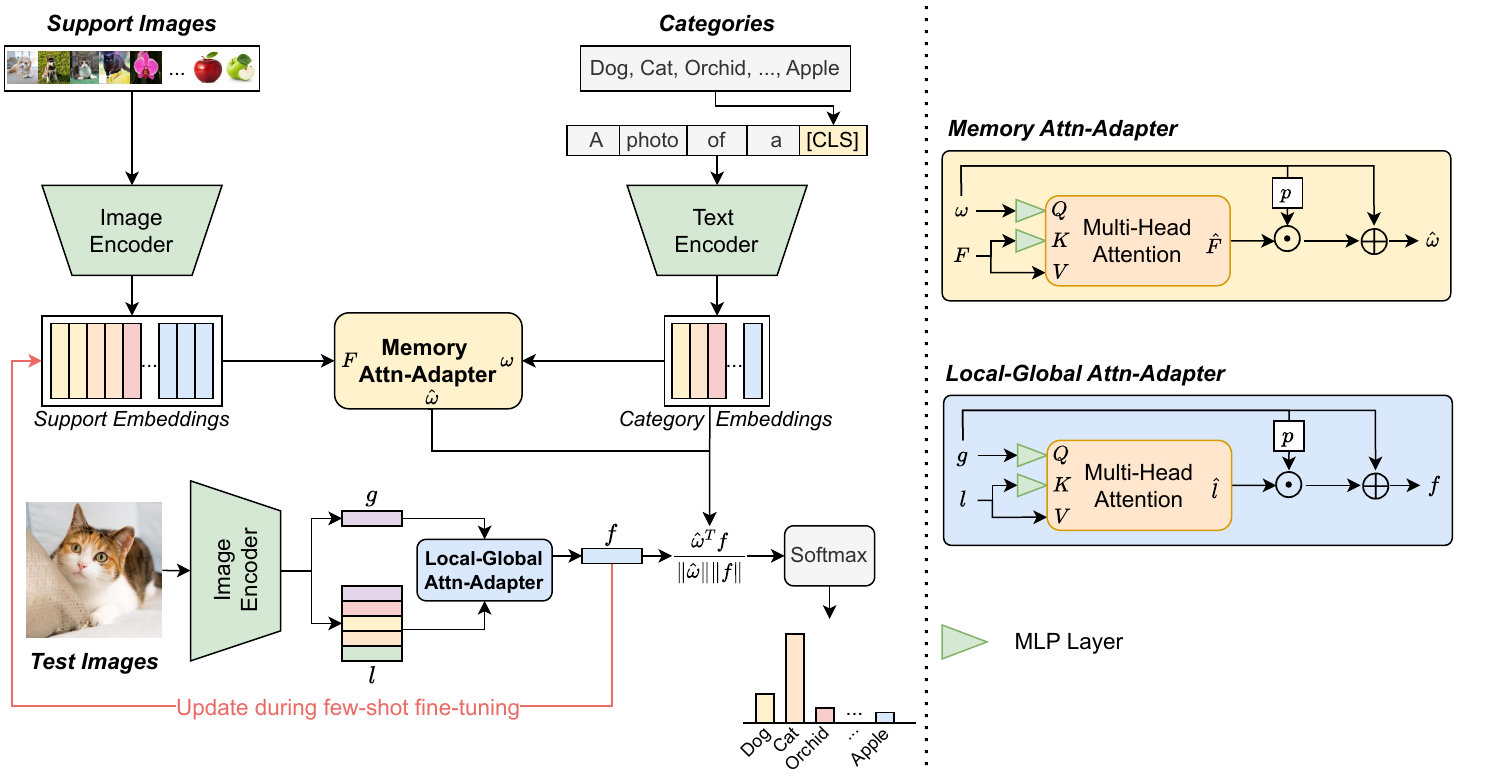}
    \caption{Illustration of the Attn-Adapter model, which utilizes a trainable network with two attention-based components to adjust the category embeddings using few-shot images as guidance.}
    \label{main}
\end{figure*}

\subsection{Revisiting CLIP, Tip- and Meta-Adapter}
\label{31}
CLIP~\cite{radford2021learning} achieves strong zero-shot performance by contrastively training on large-scale noisy image-text pairs~\cite{gu2021open, du2022learning}. For zero-shot classification, CLIP computes cosine similarities between an image feature $f \in \mathbb{R}^{D \times 1}$ and a set of text-derived class embeddings $\{w_i\}_{i=1}^N$, where $w_i \in \mathbb{R}^{D \times 1}$ and $N$ is the number of classes. Text features are generated from templates like “a photo of [CLS]”.
The predicted logits of the given image, $y$, belonging to the $i$-th class can be formulated as:
\begin{equation}
\footnotesize
\label{eq:clip1}
    \mathrm{logits}(y_c=i)=\frac{w_i^\top f}{\left \|  w_i \right \|\left \|  f \right \|},
\end{equation}

Tip-Adapter~\cite{zhang2021tip} proposes an online method for few-shot adaptation using a simple modulation function with two hyper-parameters, $\alpha$ and $\beta$, and stochastic hyper-parameter search technique. Given support images $\mathbf{x}=\{\mathbf{x}_i\}_{i=1}^N$ from $N$ classes with $K$ shots each, the predicted logits are:
\begin{equation}
\footnotesize
    \mathrm{logits}(y_c=i|\mathbf{x}, \alpha, \beta)=\frac{w_i^\top f}{\left \|  w_i \right \|\left \|  f \right \|} + \alpha \cdot \mathrm{exp}(-\beta(1 - \frac{\mathbf{F}_j^\top f}{\left \|  \mathbf{F}_j \right \|\left \|  f \right \|}))\mathbf{L}_j
\end{equation}

\noindent
Here, $\mathbf{F}_i\in \mathbb{R}^{D\times K}$ represents the embeddings of few-shot support samples, while $\mathbf{L}_i\in \mathbb{R}^{N\times K}$ denotes the corresponding one-hot labels for the $i$-th class. While effective, Tip-Adapter relies on dataset-specific hyper-parameter tuning, limiting generalization.


Meta-Adapter~\cite{song2023meta} improves Tip-Adapter’s poor generalization with a lightweight residual adapter that refines CLIP features using few-shot samples, replacing Tip-Adapter’s manual modulation. However, its reliance on zero-shot CLIP features limits dataset-specific adaptability, reducing performance on diverse tasks. Incorporating such information during training could enhance adaptability.

\subsection{Attn-Adapter}
\label{32}

In contrast with previous methods \cite{zhang2021tip, song2023meta}, which only fine-tuned the affinity matrix from few-shot support samples, Attn-Adapter proposes a new update strategy to fully leverage the trainable framework. As shown in Figure~\ref{main}, we first extract support and category embeddings using CLIP encoders. Afterward, two proposed adapters process few-shot and test samples separately. The Memory Attn-Adapter refines category embeddings by applying them as queries over support embeddings with multi-head attention. For test samples, global and local features ($g$, $l$) are passed through the Local-Global Attn-Adapter to obtain refined image embeddings $f$. The final logits are computed as:

\begin{equation}
\label{eq:clip}
    \mathrm{logits}(y_c=i|\mathbf{x})=\frac{\hat{w}_i^\top f}{\left \|  \hat{w}_i \right \|\left \|  f \right \|},
\end{equation}
where $\hat{w} = \text{Memory~Attn-Adapter}(w, \mathbf{F})$ and \\ $f = \text{Local-Global~Attn-Adapter}(g, f)$.
The $\hat{w}$ and $f$ are the refined category embeddings and image embeddings.

\subsubsection{Memory Attn-Adapter}
In the Memory Attn-Adapter, the introduced approach dynamically combines the support embeddings based on the relation between categories and few-shot images.
This approach employs a cross-attention mechanism: 
\begin{equation}
    \hat{\mathbf{F}} = \mathbf{F}^\top \sigma(\mathrm{MLP}_K(\mathbf{F})\mathrm{MLP}_Q(w)^\top /\sqrt{D}),
\end{equation}
where $\mathrm{MLP}_K$ and $\mathrm{MLP}_Q$ denote the MLP layers for key and query.
The softmax is represented by $\sigma$, $D$ is the scaling factor, and $\hat{\mathbf{F}}$ stands for the aggregated support embeddings.
Analogous to non-local filters, the Memory Attn-Adapter can ignore certain outlier samples while focusing more on samples closely aligned with the category description~\cite{xing2022class}, thus achieving robust feature representations.

Moreover, the significance of textual and visual elements for few-shot learning can differ significantly across various data distributions~\cite{gao2021clip}.
To address this, we introduce a learnable projector $p$ designed to dynamically adjust the balance between category embeddings and combined support embeddings.
As a result, the enhanced category embedding is derived as follows: $\hat{w}=w + p(w) \odot \hat{\mathbf{F}},$
where $\odot$ represents the Hadamard product.
By training with the few-shot samples, $p$ can tailor the proportion based on the category descriptions, which allows our method to successfully merge few-shot learning with zero-shot knowledge.

\subsubsection{Local-Global Attn-Adapter}
Inspired by LoCoOp \cite{miyai2024locoop} and Gallop \cite{lafon2025gallop} in using the local features to enhance the global features in prompt learning, the Memory Attn-Adapter aggregates the local and global features to generate image embeddings $f$, this $f$ is then used to update the support embeddings.
Here, we also use the cross-attention mechanism with the global features as query, and the local features as key and value.
\begin{equation}
\footnotesize
    \hat{l} = l^\top \sigma(\mathrm{MLP}_K(l)\mathrm{MLP}_Q(g)^\top /\sqrt{D}),
\end{equation}
where $\mathrm{MLP}_K$ and $\mathrm{MLP}_Q$ denote the MLP layers for key and query.
The enhanced embedding is derived as follows:
\begin{equation}
\label{eq2}
    f=g + p(g) \odot \hat{l},
\end{equation}
where $\odot$ represents the Hadamard product.

In contrast to the typical transformer encoder~\cite{vaswani2017attention}, our strategy incorporates Multilayer Perceptron (MLP) layers solely for the query and key components, thus the zero-shot value is unchanged.

\subsubsection{Training Objectives}
The objective function is a weighted combination of contrastive loss $\mathcal{L}_{ce}$ and regularization loss $\mathcal{L}_{l2}$:
   \[
   \mathcal{L} = \mathcal{L}_{ce} + \lambda \mathcal{L}_{l2},
   \]
where
\begin{align}
\footnotesize
   \mathcal{L}_{ce} &= -\sum_{\mathbf{x} \in \mathbf{X}} \log \left( \frac{\exp(d(\mathbf{x}, \hat{w}_y) / \tau)}{\sum_{i=1}^{N_c} \exp(d(\mathbf{x}, \hat{w}_i) / \tau)} \right), \\
   \mathcal{L}_{l2} &= \|f - g\|_2^2
\end{align}

   here \(d(\cdot, \cdot)\) is cosine similarity, and \(\tau\) is the temperature. 

\begin{table*}[t]\small
\begin{threeparttable}
\caption{Comparison of cross-dataset generalization based on ImageNet~\cite{deng2009imagenet} pre-training. The Tip-Adapter, Meta-Adapter, and Attn-Adapter are fine-tuned on ImageNet and frozen for other datasets. $\Delta$ reflects the improvement against the latest SOTA.}
\centering
\setlength{\tabcolsep}{1.5mm}{
\begin{tabular}{lcccccccccc|cc}
  \toprule
  Method                             & FGVC                    & Pets & SUN397 & UCF101 & Caltech101 &  DTD   & EuroSAT   & Food101   & Cars & Flowers  & Avg. & $\Delta$  \\ \midrule
  Zero-shot CLIP                      & 0.42                     & 56.25      & 28.96  & 21.05  & 60.62      & 10.00   & 4.17   & 77.40  & 55.70  & 66.00     &  38.06  & - \\
  \midrule
  
  \textcolor{gray}{Tip-Adapter*}  & \textcolor{gray}{13.96} & \textcolor{gray}{68.75}      & \textcolor{gray}{45.16}  & \textcolor{gray}{40.09}  & \textcolor{gray}{68.33}      & \textcolor{gray}{42.92}  & \textcolor{gray}{56.25}    & \textcolor{gray}{79.50} & \textcolor{gray}{75.20} & \textcolor{gray}{94.90}   & \textcolor{gray}{58.51} & \textcolor{gray}{-} \\
  
  Tip-Adapter                         & 13.96                   & 67.19      & 43.80  & 39.47  & 67.08      & 40.00  & 56.25    & 77.80  & 66.70  & 89.90 & 56.22 & \textcolor{gray}{-}  \\
  \midrule

  \textcolor{gray}{Meta-Adapter*} & \textcolor{gray}{19.58} &\textcolor{gray}{72.66}&\textcolor{gray}{51.25}&\textcolor{gray}{52.28}&\textcolor{gray}{71.46}&\textcolor{gray}{49.17}& \textcolor{gray}{64.58}   & \textcolor{gray}{81.33} & \textcolor{gray}{78.15} & \textcolor{gray}{95.12}   & \textcolor{gray}{63.56} & - \\
   
  Meta-Adapter                        &  15.21         & 72.66  &   48.54   &   47.54   &   67.92   &   48.33   & 62.50    & 79.00  & 67.30  & 93.50  & 60.25 & - \\

  \midrule
  \textcolor{gray}{Attn-Adapter*} & \textcolor{gray}{\textbf{31.25}} &   \textcolor{gray}{\textbf{74.22}}    &   \textcolor{gray}{\textbf{62.66}}    &   \textcolor{gray}{\textbf{55.44}}    &   \textcolor{gray}{\textbf{75.00}}    &   \textcolor{gray}{\textbf{60.83}}    & \textcolor{gray}{\textbf{79.19}}    & \textcolor{gray}{\textbf{84.21}} & \textcolor{gray}{\textbf{78.68}} & \textcolor{gray}{\textbf{95.65}} & \textcolor{gray}{\textbf{69.71}} & \textcolor{gray}{\textbf{\textbf{+6.15}}} \\
  
  Attn-Adapter                        &  \textbf{22.92}         & \textbf{73.97} & \textbf{55.02} & \textbf{49.93} & \textbf{68.33} & \textbf{50.67} & \textbf{66.25}    & \textbf{83.12} & \textbf{73.51} & \textbf{95.13}   & \textbf{63.89} & \textbf{+3.64} \\

  \bottomrule
\end{tabular}}
\begin{tablenotes}
 \item[*] indicates individually searching hyper-parameters or training for each evaluation dataset.
\end{tablenotes}
\label{method-table}
\end{threeparttable}
\end{table*}

\section{Experiments}

We evaluate Attn-Adapter on two tasks: cross-dataset generalization and cross-category, comparing it with Zero-shot (ZS) CLIP~\cite{radford2021learning}, Tip-Adapter~\cite{zhang2021tip}, and Meta-Adapter~\cite{song2023meta}.

\textbf{Datasets} For cross-category generalization, we use eight classification benchmarks: ImageNet~\cite{deng2009imagenet}, FGVCAircraft~\cite{maji2013fine}, OxfordPets~\cite{parkhi2012cats} (Pets), SUN397~\cite{xiao2010sun}, UCF101~\cite{soomro2012ucf101}, Caltech101~\cite{fei2004learning}, DTD~\cite{cimpoi2014describing}, and EuroSAT~\cite{helber2019eurosat}. Following~\cite{song2023meta}, categories are split into base (easy) and novel (hard) sets using zero-shot CLIP accuracy to simulate a challenging setup. For cross-dataset generalization, we further evaluate on ImageNet~\cite{deng2009imagenet} and its variants: ImageNet-A~\cite{hendrycks2021natural}, -R~\cite{hendrycks2021many}, -Sketch~\cite{wang2019learning}, and -V2~\cite{deng2009imagenet}.

\textbf{Implementation details} We use ResNet-50~\cite{he2016deep} and ViT-B/16~\cite{dosovitskiy2020image} as CLIP backbones. Following prior work~\cite{radford2021learning, zhang2021tip}, we apply prompt ensembling with seven templates. We strictly adhere the training settings in Meta-Adapter~\cite{song2023meta} including batch size, AdamW~\cite{loshchilov2017decoupled} optimizer, cosine schedule, and number of epochs on the base dataset.

\begin{table}[h]\small
\caption{Quantitative results of domain generalization experiments between Tip-Adapter, Meta-Adapter, and Attn-Adapter. The data in parentheses records the changes brought by comparing with Zero-shot CLIP. Adpt stands for Adapter.}
\label{domain}
\setlength{\tabcolsep}{1.2mm}
\centering
\begin{tabular}{cllllll}
  \toprule
  \multirow{2}{*}{Backbone} & \multirow{2}{*}{Model} & \multicolumn{4}{c}{Target Datasets}         \\   \cmidrule(r){3-7}
                            &                   & IN-A   & IN-R   & IN-S & IN-V2 & Avg \\  \midrule
  \multirow{4}{*}{RN-50}     & ZS CLIP    & 23.88        & 60.54        & 35.45            & 53.25            & 43.28 \\
                            & Tip-Adpt       & 23.25 & 58.73 & 34.77     & -     & - \\
                            & Meta-Adpt      & 23.71 & 59.96 & 35.54     & -     & - \\
                            & Attn-Adpt      & \textbf{37.61} & 60.14 & \textbf{47.88}    & \textbf{65.47}     & \textbf{52.78} \\  \midrule    
                            
  \multirow{4}{*}{ViT-B/16} & ZS CLIP    & 50.65        & 77.82        & 48.42            & 62.30            & 59.80 \\
                            & Tip-Adpt       & 49.89 & 76.94 & 48.13     & -     & - \\
                            & Meta-Adpt      & 51.12 & 77.54 & 48.76     & -     & - \\
                            & Attn-Adpt      & \textbf{62.52} & \textbf{78.71} & \textbf{59.70}    & \textbf{73.91}     & \textbf{68.71} \\  \bottomrule

\end{tabular}
\end{table}

\subsection{Cross-Dataset Generalization}
\label{42}
We evaluate cross-dataset generalization by training on ImageNet and testing on other datasets in a zero-shot manner under a 16-shot setup with frozen parameters. Table \ref{method-table} shows that Attn-Adapter outperforms other baseline methods in all datasets with an average gain of 3.64\%. In terms of training time, it matches Meta-Adapter’s efficiency while surpassing online methods, as shown in \textbf{Supplementary Material}.

We evaluated domain generalization using models trained on ImageNet, tested on ImageNet-A~\cite{hendrycks2021natural}, -R~\cite{hendrycks2021many}, -Sketch~\cite{wang2019learning}, and -V2, following~\cite{zhou2022conditional}. Tip-Adapter, with ImageNet-tuned $\alpha$ and $\beta$, underperforms Zero-shot CLIP due to overfitting (e.g., -1.81\% on ImageNet-R with RN-50). Meta-Adapter slightly improves over Zero-shot CLIP on some variants. Attn-Adapter consistently outperforms baselines, with 10\%+ gains on ImageNet-A and -Sketch, and 12\%+ on ImageNet-V2, demonstrating superior adaptability to distribution shifts for real-world applications.

\begin{table}[h]\small
\caption{Quantitative results of in-domain generalization settings on OxfordPets, UCF101, Caltech101 (Caltech), DTD, and FGVCAircraft (FGCV) datasets between Attn-Adapter and other methods.}
\label{base-to-novel-table}
\setlength{\tabcolsep}{0.5mm}
\centering
\begin{tabular}{lcccccccc}
  \toprule
  \multirow{2}{*}{Model} & \multicolumn{2}{c}{ImageNet} & \multicolumn{2}{c}{Pets} & \multicolumn{2}{c}{SUN397} & \multicolumn{2}{c}{EuroSAT}   \\   \cmidrule(r){2-9}
                         & \textcolor{gray}{Base}   & Novel             & \textcolor{gray}{Base}   & Novel                 & \textcolor{gray}{Base}   & Novel          & \textcolor{gray}{Base}   & Novel   \\  \midrule
  ZS CLIP         & \textcolor{gray}{71.81}  & 32.89             & \textcolor{gray}{92.89}  & 56.25                 & \textcolor{gray}{71.28}  & 28.96           & \textcolor{gray}{48.21}        & 4.17         \\
  Tip-Adpt            & \textcolor{gray}{74.16}  & 36.51             & \textcolor{gray}{94.83}  & 68.75                 & \textcolor{gray}{73.04}  & 45.16          & \textcolor{gray}{83.04}        & 56.25        \\
  Meta-Adpt           & \textcolor{gray}{66.08}  & 40.19             & \textcolor{gray}{92.03}  & 72.66                 & \textcolor{gray}{72.95}  & 51.25          & \textcolor{gray}{68.75}        & 64.58        \\  \midrule
  Attn-Adpt           & \textcolor{gray}{87.35}  & \textbf{54.99}             & \textcolor{gray}{92.67}  & \textbf{74.22}                 & \textcolor{gray}{77.89}  & \textbf{62.66}          & \textcolor{gray}{83.93}        & \textbf{79.19}        \\  \bottomrule
\end{tabular}
\begin{tabular}{lcccccccc}
  \toprule
  \multirow{2}{*}{Model} & \multicolumn{2}{c}{UCF101} & \multicolumn{2}{c}{Caltech} & \multicolumn{2}{c}{DTD} & \multicolumn{2}{c}{FGVC}   \\   \cmidrule(r){2-9}
                         & \textcolor{gray}{Base}   & Novel             & \textcolor{gray}{Base}   & Novel                 & \textcolor{gray}{Base}   & Novel          & \textcolor{gray}{Base}   & Novel   \\  \midrule
  ZS CLIP         & \textcolor{gray}{79.42}  & 21.05             & \textcolor{gray}{93.39}  & 60.62                 & \textcolor{gray}{59.38}  & 10.00           & \textcolor{gray}{23.84}        & 0.42         \\
  Tip-Adpt            & \textcolor{gray}{85.17}  & 40.09             & \textcolor{gray}{95.09}  & 68.33                 & \textcolor{gray}{68.36}  & 42.92          & \textcolor{gray}{30.27}        & 13.96        \\
  Meta-Adpt           & \textcolor{gray}{82.44}  & 52.28             & \textcolor{gray}{93.39}  & 71.46                 & \textcolor{gray}{64.26}  & 49.17          & \textcolor{gray}{27.32}        & 19.58        \\  \midrule
  Attn-Adpt           & \textcolor{gray}{85.36}  & \textbf{55.44}             & \textcolor{gray}{94.73}  & \textbf{75.00}                 & \textcolor{gray}{68.36}  & \textbf{60.83}          & \textcolor{gray}{31.70}        & \textbf{31.25}        \\  \bottomrule
\end{tabular}
\end{table}

\subsection{Cross-Category Generalization}
\label{41}

As shown in Table~\ref{base-to-novel-table}, Tip-Adapter excels on base datasets (e.g., 94.83\% OxfordPets, 95.09\% Caltech101) but struggles on novel categories (e.g., 40.09\% UCF101). Conversely, Attn-Adapter outperforms baselines on both, achieving 87.35\% ImageNet, 83.93\% EuroSAT (base), and 79.19\% EuroSAT, 62.66\% SUN397 (novel). Its attention-based mechanism dynamically refines embeddings, integrating global and local features for robust generalization. Please refer to \textbf{Supplementary Material} for more evaluation.

\section{Acknowledgements}
This work was supported in part by IITP grant funded by the Korean government (MSIT) under IITP-2025-RS-2020-II201821 (30\%), RS-2024-00459512 (30\%), RS-2021-II212068 (20\%), and RS-2019-II190421 (20\%).

\section{Conclusion}
We present Attn-Adapter, a lightweight online few-shot learning framework enhancing vision-language models utilizing two trainable modules: Memory Attn-Adapter and Local-Global Attn-Adapter. These components refine category and image embeddings using minimal labeled examples without backbone updates by injecting dataset-specific inductive bias during inference. Evaluated on cross-category and cross-dataset generalization, Attn-Adapter outperforms prior methods across image classification benchmarks with low inference cost, generalizing well across backbones (ResNet, ViT). Furthermore, Attn-Adapter generalizes well across backbones (e.g., ResNet, ViT) and is particularly effective in low-shot and domain-shift settings. In future work, we plan to extend Attn-Adapter to more complex downstream tasks such as open-vocabulary object detection and semantic segmentation, further validating its generality and flexibility in broader vision-language applications.
\section{Supplementary Materials}
\paragraph{Computational Complexity} Additionally, Table~\ref{time} compares training times on ImageNet using a single GeForce RTX 3090, revealing Attn-Adapter’s efficiency as only introducing 5\% overhead compared to Meta-Adapter, solidifying Attn-Adapter’s practical edge in real-world deployment.

\begin{table}[h]\small
\centering
\begin{threeparttable}
\caption{Training time on ImageNet of Attn-Adapter and other methods on a single GeForce RTX 3090.}
\label{time}
\centering
\setlength{\tabcolsep}{0.2mm}{
\begin{tabular}{ccccc}
  \toprule
  CLIP-Adapter   &  CoOp       & CoCoOp  & Meta-Adapter  & Attn-Adapter  \\
  200 epochs   &  200 epochs        & 10 epochs & 10 epochs  & 10 epochs  \\ \midrule
  17h 30min           & 15h 10min       &  23h 30min    & \textbf{20min}        & 21min    \\
  \bottomrule
\end{tabular}}
\end{threeparttable}
\end{table}

\begin{table*}[h]\small
\caption{Quantitative results on ImageNet of different models utilized various vision backbones.}
\setlength{\tabcolsep}{3mm}
\centering
\begin{tabular}{lcccccc}
  \toprule
 Model   & RN50 & RN101 & ViT-B/32 & ViT-B/16  & RN50×16 & RN50×64  \\  
\midrule
  Zero-shot CLIP         & 32.82  & 39.22             & 40.10  & 45.77  & 50.10  & 54.67          \\
  Tip-Adapter            & 36.51  & 42.42             & 43.71  & 49.84  & 53.08  & 57.99         \\
  Meta-Adapter           & 40.19  & 47.01             & 46.91  & 52.60  & 55.51  & 60.41         \\  \midrule
  Attn-Adapter           & \textbf{50.35}  & \textbf{58.17}             & \textbf{56.42}  & \textbf{55.65}  & \textbf{63.69}  & \textbf{66.27}         \\  \bottomrule
\end{tabular}
\label{different backbone}
\end{table*}

\paragraph{Backbone Compatibility} To effectively validate Attn-Adapter’s capabilities, we conducted experiments on ImageNet using a variety of visual encoders: ResNet-50, ResNet-101, ViT-B/32, ViT-B/16, RN50×16, and RN50×64, each offering different architectural strengths and complexities. As shown in Table~\ref{different backbone}, Meta-Adapter consistently outperforms Tip-Adapter across all backbones, benefiting from its meta-learning framework that optimizes for adaptability. However, Attn-Adapter surpasses both, achieving remarkable gains, particularly with advanced backbones like ViT-B/16 (55.65\% vs. Meta-Adapter’s 52.60\%) and RN50×64 (66.27\% vs. 60.41\%). This improvement is driven by Attn-Adapter’s ability to exploit the enhanced representational power of larger models, such as ViT’s attention-based processing or RN50×64’s deeper convolutions, which provide richer feature extraction. The consistent upward trend across backbones suggests that Attn-Adapter scales effectively with architectural advancements, a promising attribute for future integration with vision-language models. Unlike Tip-Adapter, which plateaus due to its hyper-parameter constraints, Attn-Adapter mitigates over-fitting by dynamically adapting its embeddings, achieving state-of-the-art accuracy on novel datasets and positioning it as a highly scalable solution for diverse applications.

\paragraph{Comparison with Offline Methods}

\begin{table*}
\caption{Comparison of Zero-Shot CLIP, CLIP-Adapter, CoOp, CoCoOp, Meta-Adapter, and Attn-Adapter on ImageNet, UCF101, Caltech101, DTD, and FGVCAircraft in an in-domain generalization setting. H: Harmonic mean, highlighting the trade-off between base and novel class performance \cite{zhou2022conditional}.}
\label{base-to-novel-table-2}
\centering
\resizebox{\textwidth}{!}{
\begin{tabular}{lccccccccccccccc}
  \toprule
  Dataset & \multicolumn{3}{c}{ImageNet} & \multicolumn{3}{c}{UCF101} & \multicolumn{3}{c}{Caltech101} & \multicolumn{3}{c}{DTD} & \multicolumn{3}{c}{FGVCAircraft}  \\   \hline
    Model    & Base   & Novel   & H     & Base   & Novel   & H      & Base   & Novel    &  H      & Base   & Novel   & H      & Base   & Novel  &  H      \\  \midrule
  Zero-shot CLIP         & 71.9   & 32.8    & 45.0  & 79.4   & 21.1    & 33.4   & 95.4   & 60.6     & 74.1   & 59.3   & 8.2     & 14.3   & 23.9   & 0.6    & 1.2    \\
  CLIP-Adapter           & 76.3   & 15.1    & 25.3  & 89.4   & 5.4     & 10.2   & 97.3   & 39.3     & 54.0   & 70.2   & 2.0     & 3.9    & 32.1   & 0.3    & 0.6    \\
  CoOp                   & 75.3   & 2.7     & 5.2   & 89.3   & 1.0     & 2.0    & 97.2   & 31.3     & 47.4   & 71.6   & 1.5     & 2.9    & 32.7   & 0.3    & 0.6    \\
  CoCoOp                 & 75.5   & 33.9    & 46.8  & 86.5   & 9.1     & 16.5   & 96.8   & 60.9     & 74.8   & 69.1   & 3.0     & 5.8    & 30.0   & 0.8    & 1.6    \\
  Meta-Adapter           & 76.3   & 40.8    & 53.2  & 82.4   & 47.7    & 60.4   & 94.9   & 76.1     & 84.4   & 64.1   & 49.1    & 55.6   & 30.8   & 17.1   & 21.9   \\
  Attn-Adapter           & \textbf{81.4} & \textbf{43.6} & \textbf{56.8} & \textbf{85.4} & \textbf{52.2} & \textbf{64.8} & \textbf{95.7} & \textbf{79.3} & \textbf{86.7} & \textbf{68.5} & \textbf{60.8} & \textbf{64.4} & \textbf{33.7} & \textbf{31.1} & \textbf{32.3} \\  \bottomrule
\end{tabular}}
\end{table*}

Table~\ref{base-to-novel-table-2} presents ablation studies comparing Attn-Adapter against offline methods, including CLIP-Adapter, CoOp, CoCoOp, and Meta-Adapter \cite{zhou2022conditional, zhou2022learning, gao2021clip, song2023meta}, using CoCoOp’s base-to-novel generalization framework. Attn-Adapter outperforms all methods, particularly on novel classes, achieving 43.6\% accuracy on ImageNet (a 2.8\% gain over Meta-Adapter’s 40.8\%) and 52.2\% on UCF101 (versus 47.7\%). The harmonic mean (H), which balances base and novel class performance, further highlights Attn-Adapter’s superiority, with scores of 56.8 on ImageNet and 64.8 on UCF101 compared to Meta-Adapter’s 53.2 and 60.4, respectively. This superior effect is clearly visible on FGVCAircraft. Unlike CoOp, which prioritizes base classes (75.3\% on ImageNet) at the expense of novel ones (2.7\%), Attn-Adapter maintains robust performance across both, demonstrating enhanced adaptability. These results position Attn-Adapter as a highly effective solution for applications requiring generalization to new categories while retaining prior knowledge.

{
    \small
    \bibliographystyle{ieeenat_fullname}
    \bibliography{main}
}

\end{document}